\documentclass[11pt,a4paper]{article}
\usepackage[hyperref]{acl2021}
\usepackage{times}
\usepackage{latexsym}

\usepackage{microtype}

\aclfinalcopy 
\def\aclpaperid{1857} 

\usepackage{times}
\usepackage{latexsym}

\usepackage{amsmath}
\usepackage{amssymb}
\usepackage{todonotes}
\usepackage{algpseudocode}
\usepackage{algorithm}
\usepackage{multirow}
\usepackage{booktabs} 

\newcommand{\mbf}[1]{\mathbf #1}
\newcommand{\mc}[1]{\mathcal #1}
\newcommand{\eat}[1]{}

\DeclareMathOperator*{\argmax}{\arg\!\max}
\DeclareMathOperator*{\argmin}{\arg\!\min}
\newcommand{\citeay}[1]{\citeauthor{#1}~(\citeyear{#1})}
\newcommand{\ra}[2]{#1$\rightarrow$#2}

\usepackage{microtype}

\title{Energy-Based Reranking: Improving Neural Machine Translation\\ Using Energy-Based Models}

\author{Sumanta Bhattacharyya, Amirmohammad Rooshenas\thanks{ $ $ Amirmohammad Rooshenas is the corresponding author.} 
\\Department of Computer Science,
College of Computing and Informatics\\
University of North Carolina Charlotte\\
\texttt{\{sbhatta9,rooshenas\}@uncc.edu}
\vspace{0.1in}

\\\textbf{Subhajit Naskar, Simeng Sun, Mohit Iyyer, and Andrew McCallum} \\
  College of Information and Computer Science,
  University of Massachusetts Amherst \\
  \texttt{\{snaskar,simeng,miyyer,mccallum\}@cs.umass.edu}
}

\date{}

\begin{document}
\maketitle

\begin{abstract}
    The discrepancy between maximum likelihood estimation (MLE) and task measures such as BLEU score has been studied before for autoregressive neural machine translation (NMT) and resulted in alternative training algorithms~\cite{ranzato&al16, norouzi&al16, shen&al16,wu&al18}. However, MLE training remains the de facto approach for autoregressive NMT because of its computational efficiency and stability. 
    Despite this mismatch between the training objective and task measure, we notice that the samples drawn from an MLE-based trained NMT support the desired distribution -- there are samples with much higher BLEU score comparing to the beam decoding output. To benefit from this observation, we train an energy-based model to mimic the behavior of the task measure (i.e., the energy-based model assigns lower energy to samples with higher BLEU score), which is resulted in a re-ranking algorithm based on the samples drawn from NMT: energy-based re-ranking (EBR). We use both marginal energy models (over target sentence) and joint energy models (over both source and target sentences). Our EBR with the joint energy model consistently improves the performance of the Transformer-based NMT: +3.7 BLEU points on IWSLT'14 German-English, +3.37 BELU points on Sinhala-English, +1.4 BLEU points on WMT'16 English-German tasks.
\end{abstract}

\section{Introduction}

Autoregressive models are widely used for neural machine translation (NMT)~\cite{bahdanau&al15,gehring&al17a,vaswani&al17}. The autoregressive factorization provides a tractable likelihood computation as well as efficient sampling. The former results in the effective maximum likelihood estimation (MLE) for training the parameters of NMT models. However, optimizing likelihood does not guarantee an improvement in task-based measures such as the BLEU score, which has motivated directly optimizing task measures with reinforcement learning~\cite{ranzato&al16, norouzi&al16, shen&al16,bahdanau&al17,wu&al18}. However, for NMT, these training algorithms are often used in conjunction with MLE training~\cite{wu&al18} or as fine-tuning~\cite{choshen&al20}.




\begin{figure*}
    \centering
    \begin{tabular}{c c}
          \includegraphics[width=0.7\columnwidth]{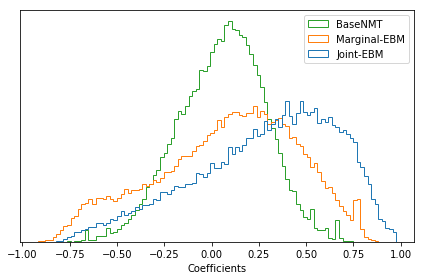}& 
          \includegraphics[width=0.7\columnwidth]{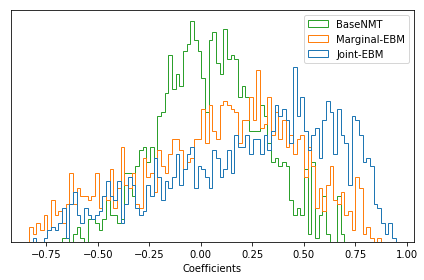} \\
    \end{tabular}

    \vspace{-0.1in}
    \caption{Distribution of the Spearman rank-order correlation coefficients for the training data (left) and test data (right) of the IWSLT'14 German-English task.}
    \label{fig:spearman}
    \vspace{-0.2in}
\end{figure*}

Interestingly, we observe that samples drawn from an NMT model trained using MLE may have higher quality (measured with BLEU) than the outputs of beam search.  In particular, we draw 100 target samples for each source sentence  from an NMT model trained using MLE on the IWSLT'14 German-English task, and observe that an oracle ranker -- i.e. $\argmax_{\mbf{y} \sim P_{\text{NMT}}(\mbf{y}|\mbf{x})} \text{BLEU}(., \mbf{y}^*)$, where $(\mbf{x}, \mbf{y}^*)$ is the pair of source and gold target sentence -- achieves the high score of 67.54, while the beam decoding achieves 33.87. We also look at the distribution of the Spearman rank correlation coefficient of the drawn samples with respect to the log probability score of the baseline NMT (BaseNMT). Figure~\ref{fig:spearman} shows that there is no strong correlation between the BLEU score ranking of samples and the log probability score ranking for the majority of source sentences; thus, maximum a priori (MAP) decoding is incapable of finding the desired output. In parallel to our study, \citeay{eikema&aziz20} also report that the mismatch regarding MLE training of autoregressive models is attributable to the distribution of the probability mass rather than the parameter estimation, resulting in a poor MAP decoding.  

Instead of looking for an alternate algorithm for parameter estimation, these results motivate us to explore training a parametric approximation of the metric, here BLEU score: $\omega_\theta(\mathbf{y}, \mbf{x}) \approx \text{BLEU}(\mbf{y}, \mbf{y}^*)$. Therefore the decoding becomes: $\argmax_{\mbf{y} \sim P_{\text{NMT}}(.|\mbf{x})} \omega_\theta(\mathbf{y},\mbf{x})$. 

We use energy-based models (EBMs) to parameterize $\omega_\theta(\mathbf{y},\mbf{x})$. EBMs~\cite{lecun&al06} are general parametric models that assign a scalar energy value to each configuration of input variables, thus defining an unnormalized probability distribution. Although computing the partition function is intractable for general EBMs, we only require the relative energy of the sampled sentences from the BaseNMT model, thus canceling out the normalization constant. In this paper we use two different energy-based models: marginal energy model (Marginal-EBM) defined only over target sentences and joint energy model (Joint-EBM) defined over both source and target sentences.

Figure~\ref{fig:spearman} also shows the correlation coefficient of the energy ranking and BLEU score using both Marginal-EBM and Joint-EBM. The shift in the coefficient distribution suggests that decoding based on energy scores results in better BLEU scores compared to decoding based on the log probability scores of the BaseNMT model. Also we observe that Joint-EBM works better than using Marginal-EBM as Joint-EBM better captures the correlation of source and target sentences, while Marginal-EBM is not directly conditioned on the source sentence. 

In this paper, we describe how to train EBMs\footnote{The code is available at \url{https://github.com/rooshenas/ebr_mt}} to achieve the desired ranking. Our energy ranker consistently improves the performance of Transformer-based NMT on German-English, Romanian-English and Italian-English tasks from IWSLT'14, the French-English task from IWSLT'17, German-English task from WMT'14, and English-German task from WMT'16, as well as the low-resource Sinhala-English and Nepali-English tasks described in the FLoRes dataset~\cite{guzman&al19}. 








\section{Energy-Based Reranking}

\begin{figure}
    \centering
    \includegraphics[width=\columnwidth]{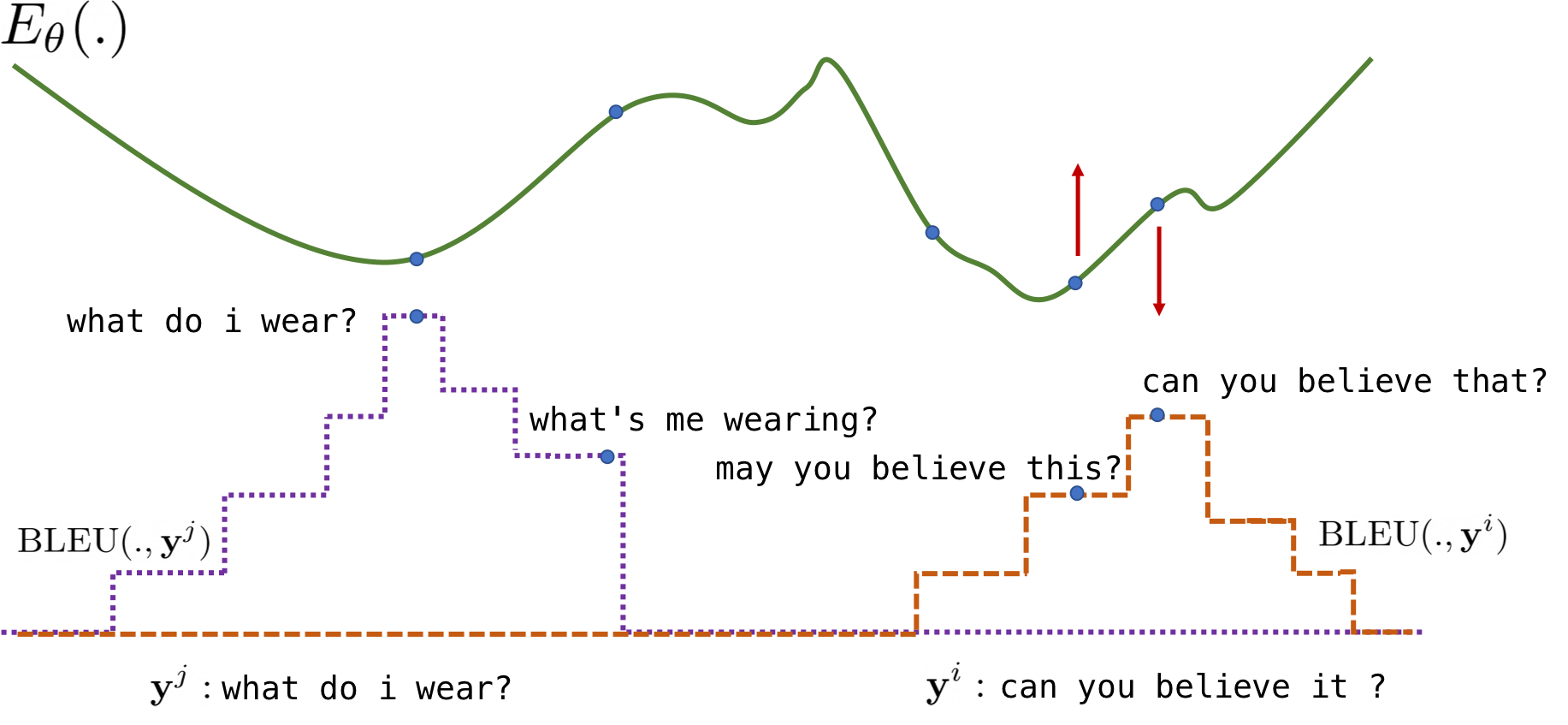}
    \caption{The EBM is trained such that its energy landscape is consistent with the BLEU score. Marginal-EBM is not conditioned on the source sentence, thus each local region is trained to have similar ranking as that BLEU score for the samples in the region.}
    \label{fig:energy}
\end{figure}


Using EBM $E_\theta$ to reweight the samples from an NMT defines a new probability distribution over the output sentences (see \citeay{grover&al19}): 
\(P_\theta(\mathbf{y}|\mbf{x}) \propto  P_\text{NMT} (\mathbf{y}|\mathbf{x}) \exp(\frac{-E_\theta(\mathbf{y},\mbf{x})}{T})\),
where $T$ is temperature.
The ideal re-ranker requires an EBM with the energy function $E_\theta(\mbf{y},\mbf{x})$ such that $P_\theta(\mbf{y}|\mbf{x})$ and $\text{BLEU}(\mbf{y}, \mbf{y}^i)$ have similar modes for all  $(\mbf{x}^i,\mbf{y}^i)\in \mc{D}$, where $\mc{D}$ is an empirical data distribution. 
To train $\theta$ we use rank-based training~\cite{rohanimanesh&al11, rooshenas&al18,rooshenas&al19}. Rank-based training enforces that the samples from $P_\theta(.)$ have similar ranking with respect to both the energy score and task measure (see Figure~\ref{fig:energy}).




To sample from $P_\theta(\mbf{y}|\mbf{x})$, we sample $k$ sentences from $P_\text{NMT}(\mbf{y}|\mbf{x})$ using multinomial sampling from locally normalized distributions over the output and reweight the samples based on the energy network $\exp(\frac{-E_\theta(\mbf{y},\mbf{x})}{T})$.
Then we resample two sentences, $\mbf{y}_1$ and $\mbf{y}_2$, from the renormalized set, which defines a conditional distribution: \( P^i(\mbf{y}|\mbf{x}) = \frac{\exp(-E_\theta(\mbf{y},\mbf{x})/T)}{\sum_k \exp(-E_\theta(\mbf{y}_k,\mbf{x})/T)} \) (a similar sampling approach has been used in \citeay{deng&al20}). 
Now we train the energy model such that the ranking of $\mbf{y}_1$ and $\mbf{y}_2$ with respect to the energy model is consistent with their ranking with respect to the task metric, BLEU score. 


In general, we assume $\mbf{y}_h$ is the sentence with the higher BLEU score and $\mbf{y}_l$ is the sentence with with the lower BLEU score. Therefore, the training objective of $E_\theta(\mbf{y},\mbf{x})$ becomes:
\begin{align}
&M = \alpha (\text{BLEU}(\mbf{y}_h, \mbf{y}_i) - \text{BLEU}(\mbf{y}_l, \mbf{y}_i)) \nonumber \\
&\xi(\mbf{y}_i, \mbf{x}_i) = M + E_\theta(\mbf{y_h},\mbf{x}_i) - E_\theta(\mbf{y}_l,\mbf{x}_i) \nonumber \\
&\min_\theta \sum_{(\mbf{y}_i,\mbf{x}_i)\in \mc{D}}\ \max(\xi(\mbf{y}_i,\mbf{x}_i),0).
\end{align}
Where $\xi(\mbf{y}_i, \mbf{x}_i)$ is the margin violation and $\alpha$ is the margin weight. Algorithm~\ref{alg:rank} outlines the whole training procedure.


If we define the energy only over sentences of the target language, $E_\theta(\mathbf{y})$, we can share the energy-model among multiple language pairs with the same target language. In this case we have to, first, sample the language $l$ from our language set and then sample a sentence pair from the selected language training set $\mc{D}_l$. The probability of selecting a language is proportional to the number of sentences in its training set.




\begin{algorithm}
\caption{Rank-Based Training of EBM}
\small{
\begin{algorithmic}
\State $P_\text{NMT}(y|x)$ $\gets$ Pretrained NMT 
\State $E_\theta (\mbf{y},\mbf{x}) \gets$ Energy based models for target sentences
\Repeat
\State $\mc{L} \gets 0$.   
\For {batch size}  
\State Sample $(\mbf{x}_i,\mbf{y}_i)$  from $\mc{D}$
\State $Y_i \gets$ collect $k$ samples from $P_\text{NMT}(.|\mbf{x}_i)$
\State $P^i(\mbf{y}) \gets \frac{\exp(-E_\theta(\mbf{y},\mbf{x})/T)}{\sum_{\mbf{y}\in Y_i} \exp(-E_\theta(\mbf{y},\mbf{x})/T)}$  for $\mbf{y}\in Y_i$
\State $\mbf{y}_1,\mbf{y}_2 \gets$ samples from $P_i(\mbf{y})$
\State $\mbf{y}_h \gets \argmax_{\mbf{y}_1,\mbf{y}_2} \{{\tiny\text{BLEU}}(\mbf{y}_1, \mbf{y}_i), {\tiny \text{BLEU}}(\mbf{y}_2, \mbf{y}_i)\}$
\State $\mbf{y}_l \gets \argmin_{\mbf{y}_1,\mbf{y}_2} \{{\tiny \text{BLEU}}(\mbf{y}_1, \mbf{y}_i), {\tiny \text{BLEU}}(\mbf{y}_2, \mbf{y}_i)\}$
\State $M \gets \alpha ({\tiny\text{BLEU}}(\mbf{y}_h, \mbf{y}_i) - {\tiny\text{BLEU}}(\mbf{y}_l, \mbf{y}_i))$
\State $\mc{L} \gets \mc{L} + \max(M + E_\theta(\mbf{y}_h,\mbf{x}_i) - E_\theta(\mbf{y}_l,\mbf{x}_i),0)$
\EndFor
\State $\theta \gets \theta - \lambda \nabla_\theta \mc{L}$ $ \quad $ // $\lambda$ is learning rate
\Until{Convergence}
\end{algorithmic} 
}
\label{alg:rank}

\end{algorithm}

In this paper, we use BERT~\cite{devlin-etal-2019-bert} to parameterize both $E_\theta(\mbf y, \mbf x)$ and $E_\theta(\mbf y)$. Section~\ref{sec:marginal-ebm} and \ref{sec:joint-ebm} discuss the construction of $E_\theta$ in detail.

\section{Related Work}

\citeay{grover&al19} show that importance weights can be used to make generative models better fit the desired data distribution:
{$p_\theta(\mathbf{y}) \propto q(\mbf{y})\omega_\theta(\mbf{y})$},
where $q(\mbf{y})$ is a generative model that we can efficiently take samples from and $\omega_\theta(\mbf{y})$ is the importance weight function. The importance weights can be determined using a discriminator that differentiates the generated samples from the target data.  \citeay{rosenfeld&al01,parshakova&al19} define $q(\mbf{y})$ as autoregressive model and $\omega_\theta(\mbf{y})$ using a log-linear model: {$\omega_\theta(\mbf{y}) = \exp (\theta^T \phi(\mbf y))$}, where $\phi(\mbf{y})$ is the vector of sufficient statistics (features) evaluated at $\mbf{y}$. The log-linear model simplifies training the parameters $\theta$: {$\nabla_\theta p_\theta(\mbf{y}) = \sum_{\mbf{y}\in \mc{D}} \phi(\mbf{y}) -  \mathbb{E}_{\hat{\mbf{y}}\sim p_\theta(.)} \phi(\hat{\mbf{y}})$}.
The expectation term can be estimated using rejecting sampling or importance sampling given the proposal distribution $q$.
\citeay{deng&al20} extend this approach for text generation by using unrestricted EBMs instead of log-linear models: { $\omega_\theta(\mbf{y}) = \exp(-E_\theta(\mbf{y}))$}. They train the EBM using noise contrastive estimation~\cite{gutmann&hyvarinen10}. We find this less suitable for re-ranking in the translation tasks (see Section 4).

Discriminative re-ranking was first introduced by \citeay{shen&al04} for improving the performance of machine translation (MT). They have trained a linear separator using the perceptron learning algorithm to distinguish the top $r$ translations from the rest of the translations in the n-best possible outputs. The features for the discriminator are extracted from both source and target sentences. \citeay{mizumoto&matsumoto16} combine the score of MT and the linear model using more complex syntactical features to re-rank the target sentences. Here, we rely on the features learned by BERT, and given the high capacity of the energy model, we train the energy model to respect the ranking of every pair of samples. 

\citeay{gulcehre&al17} describe using language model (LM) to improve the performance of NMT using shallow and deep fusion. Shallow models combine the marginal probability of predicting each word in NMT and LM: $\log P_{\text{NMT}}(y_i|y_{<i}) + \lambda \log P_{\text{LM}}(y_i|y_{<i})$, while deep fusion concatenates the hidden states of two models before predicting each word and uses parallel data to fine-tune the weights.  Similar to deep fusion, \citeay{domhan&hieber17} feed the unnormalized output of LM to the decoder of NMT. \citeay{domhan&hieber17}  jointly train the LM and NMT using monolingual target-side data and parallel data, respectively. \citeay{sennrich&al16} augment the parallel training data with monolingual data with the target language and back-translation. 

Re-ranking with LM has also been explored by \citeay{ng&al19}, where they decode the output based on $\log p(y|x) + \lambda_1 \log p(x|y) + \lambda_2 \log p(y)$, where $p(y|x)$ is the direct model provided by NMT, $p(x|y)$ is computed via back-translation and $p(y)$ is an LM.
Our approach differs from the previous methods that use LMs for re-ranking as we train our energy-based model to be consistent with the task measure instead of using pre-trained LMs.
In our experiments, we only explore the effect of using the direct model plus LM, nevertheless, back-translation can also be added into our model for further improvement.

Recently, \citeay{salazar&al20} use masked language models (MLM) such as BERT to score hypotheses from NMT. \citeay{salazar&al20} describe the score of a MLM as pseudo-log-likelihood score (PLL). To calculate PLL score of a sentence, each token $w_i$ in the sentence is sequentially masked, which allows the calculation of $\log p(w_i| \mathbf{w}_{\backslash i})$ from the output of the MLM. The normalized pseudo-log-probability of the sentence is the average of log-probability of the masked words given the rest of the words in the sentence: $\frac{1}{N} \sum_{i=1}^N \log p(w_i|\mathbf{w}_{\backslash i})$, where $N$ is the length of the sentence. We use this approach as one of our baselines.

In parallel to our work, \citeay{guo&al2020} proposes using two different BERT models as an encoder of the source language (X-BERT) and a decoder of the target language (Y-BERT).   \citeay{guo&al2020}  add an extra trainable encoder-decoder adaption module followed by a feed-forward module to each layer of the decoder and a feed-forward module to each layer of the encoder. (Please see \citeay{guo&al2020} for more detail on the architecture.) 
 For fine-tuning XY-BERT for translation tasks, \citeay{guo&al2020} keep all XY-BERT's parameters fixed except the parameters of the new modules, and use mask-predict decoding~\cite{ghazvininejad&al19} for running test-time inference.
 \citeay{guo&al2020} report a significant improvement over prior non-autoregressive models and superior performance comparing to autoregressive methods on IWSLT'14 German-English task. Their finding is consistent with our improvement using the pretained BERT model. However, our Joint-EBM model is a different way of using BERT for translation, which does not require separate BERT models for source and target language. Please see Section~\ref{sec:xybert} for a detailed comparison. 

Finally, other works also discuss using BERT to improve the performance of NMT. \citeay{clinchant&al19} describe initializing the embedding or the whole encoder with BERT's parameters. \citeay{zhu&20} use an attention model to incorporate the output of BERT into encoder and decoder of NMT. In our approach, we use BERT as an external energy-based ranker.

\section{Experiments}





\begin{table*}[h]
\centering

\caption{BLEU score comparison for IWSLT, FLoRes, and WMT (indicated using *) tasks.}
\vspace{-0.1in}
\small{
\begin{tabular}{ l c c c c c c c c c } 
\toprule
 & De$\xrightarrow{}$En & Fr$\xrightarrow{}$En & It$\xrightarrow{}$En & Ro$\xrightarrow{}$En & Si$\xrightarrow{}$En & Ne$\xrightarrow{}$En & En$\xrightarrow{}$De & \ra{De}{En}* & \ra{En}{De}* \\
\midrule
BaseNMT + Beam & 33.87 & 31.50 & 32.08 & 33.21 & 7.10 & 6.07 & 28.83  & 30.13 & 28.84  \\ 
BaseNMT + Sample & 33.98 & 31.59 & 32.22 & 33.64 & 7.19 & 6.44 & 28.85& 30.28 &28.89  \\
BaseNMT + LM & 34.25 & 31.56 & 32.52 & 33.01 & 7.11 & 6.02 & 28.91 &  30.31&28.93      \\  
BaseNMT + MLM & 34.42 & 32.13 & 33.68 & 33.85 & 7.70 & 7.21 & 30.12 & 30.61&28.98 \\  
NCE-EBR & 34.47 & 32.00 & 32.89 & 32.23 & 7.98 & 7.36 & 28.22 & 31.42   &29.03\\ 
            Marginal-EBR & 35.68 & 33.77 & 34.00 & 34.48 & 8.62 & 7.26 & 30.82 & 31.65&29.14\\ 
Shared-EBR & 35.75 & 33.80 & 34.14 & 34.65 & 10.29 & 9.25 & - & - & -              \\
Conditional-EBM & \textbf{37.58}&\textbf{35.02} &\textbf{36.05} &\textbf{37.19} &\textbf{10.47} & \textbf{9.82} & \textbf{30.97} &\textbf{32.21}&\textbf{30.23}  \\
\midrule
Oracle & 67.54 & 68.43 & 71.77 & 73.95 & 14.71 & 11.91 & 52.14 &50.89&45.15\\
\bottomrule 
\end{tabular}
}
\label{tab:main}
\vspace{-0.1in}
\end{table*}

\begin{table}[t]
    \centering
    
    \caption{Shared-EBR performance for \ra{Si}{En} by training with difference sets of language pairs.}
    \vspace{-0.1in}
    \small{
\begin{tabular}{ c c c c c} 
\toprule
BaseNMT & + \ra{Si}{En} & + \ra{De}{En} & + \ra{Fr}{En} & all\\
\midrule
7.10 & 8.62 & 9.30 & 9.76 & \textbf{10.29} \\
\bottomrule
\end{tabular}
}
\label{tab:si}
\vspace{-0.1in}
\end{table}


\subsection{Datasets}
We use German-English (De$\rightarrow$En), Romanian-English (Ro$\rightarrow$En) and Italian-English (It$\rightarrow$En) from IWSLT'14 datasets and French-English (Fr$\rightarrow$En) from IWSLT'17 translation tasks.  We also use IWSLT'14 English-German (En$\rightarrow$De) to show that the proposed method can be expanded to translation tasks with a different target language. 
All sentences were preprocessed using byte-pair-encoding~\cite{sennrich-etal-2016-neural}. For all language pairs in IWSLT'14 and IWSLT'17, we merge the test datasets tst2010, tst2011, tst2012 and report BLEU on the merged dataset. 
We also use German-English (De$\rightarrow$En) from the WMT'14 and English-German (En$\rightarrow$De) from WMT'16 translation tasks. 

Finally, we use low-resource translation tasks Nepali-English (Ne$\rightarrow$En) and Sinhala-English (Si$\rightarrow$En) from FLoRes~\cite{DBLP:journals/corr/abs-1902-01382} translation tasks. We follow dataset distribution and preprocessing steps described in \citeauthor{DBLP:journals/corr/abs-1902-01382}~(\citeyear{DBLP:journals/corr/abs-1902-01382}) using the FLoRes implementation. FLoRes dataset contains development (dev), devtest and test dataset for both language pairs. Similar to \citeauthor{DBLP:journals/corr/abs-1902-01382}~(\citeyear{DBLP:journals/corr/abs-1902-01382}) we use the devtest dataset for all our evaluations.

\subsection{Base Model} 

We use the Transformer\footnote{We use the implementation in Opennmt~\cite{opennmt} and Fairseq~\cite{ott2019fairseq} toolkits.}\cite{vaswani&al17} as our BaseNMT.
Our Transformer architecture includes six encoder and six decoder layers, and the number of attention heads, embedding dimension and inner-layer dimension are 8, 512
and 4096, respectively. We use dropout, weight decay, label smoothing to regularize our models. We use layer normalization and early stopping. Models are optimized using Adam~\cite{kingma2014method} with parameters $\beta_1 = 0.9$, $\beta_2 = 0.98$, and $\epsilon = 1e^{-8}$ and we use the same learning rate scheduler as \citeauthor{ott2019fairseq}~(\citeyear{ott2019fairseq}). We trained our models on 1 Nvidia TITANX GPU.

\subsection{Marginal-EBM}\label{sec:marginal-ebm}
To construct the energy network over the sentences of the target language, we use a pretrained BERT~\cite{devlin-etal-2019-bert} from Huggingface~\cite{Wolf2019HuggingFacesTS} as our pretrained language model and project the hidden state of BERT for each output token into a scalar value and define the energy value of the target sentence as the average of the scalar values. We use the \textit{BERT-base uncased} model with 12 encoder layers, 768 hidden state dimension, 12 attention heads and 110M parameters. For the projection layer, we use a 2-layer MLP with 256 hidden variables. In our experiments, we only train the parameters of the projection layer and the rest of BERT's parameters remain frozen. 
We use margin weight of $\alpha=10$ and temperature $T=1000$ for our experiments. We regularize the projection layer using L2 regularization.  Models are optimized using Adam \cite{kingma2014method} with parameters $\beta_1 = 0.9$, $\beta_2 = 0.98$, and $\epsilon = 1e^{-8}$ and a learning rate of $0.01$.
We run all experiments on 1 Nvidia TESLA M40 GPU.

\subsection{Joint-EBM}\label{sec:joint-ebm}
Joint-EBM must assign a score to a pair of sentences from source and target languages, so to construct the Joint-EBM, similar to Marginal-EBM, we need a Joint-BERT. We feed the sentence pairs from source and target languages jointly to BERT, thus the name Joint-BERT. Since Joint-BERT has not been pre-trained to accept pairs of sentences from two different languages, we fine-tune it for 12 epochs using the input format of {\small \verb+[CLS]Source[SEP]Target[SEP]+} with the pairs of source and target sentences for each translation task. 
For fine-tuning, we only mask the tokens of the target sentence. For all translation tasks we use the BERT-Base, Multilingual Cased model with 12 encoder layers, 768 hidden state dimension, 12 attention heads and 110M parameters. 
After fine-tuning Joint-BERT, we follow the same architecture as Marginal-EBM for the Joint-EBM. 

\subsection{Methods}
 As the main baseline, we run beam decoding with a beam size of five over the trained BaseNMT (BaseNMT+Beam). We also use the samples drawn from the BaseNMT and report the BLEU score of the sample with the highest log-probability score on BaseNMT (BaseNMT+Sample). For all methods we use 100 target samples for each source sentence.
BaseNMT+LM draws samples from the BaseNMT and uses {$\log P_{\text{NMT}}(\mbf y|\mbf x) + \lambda \log P_{LM}(\mbf{y})$} to rank the samples ($\lambda=0.01$ out of the set of \{0.001, 0.01, 0.1\} results in the best performance). 

In our BaseNMT+LM baseline, we use pretrained language model to calculate $\log P_{LM}(\mbf{y})$. For the \{De, Fr, It, Ro, Si, Ne\}$\xrightarrow{}$En tasks, we use a pretrained Transformer-XL~\cite{DBLP:journals/corr/abs-1901-02860} \textit{transfo-xl-wt103} and for the En$\xrightarrow{}$De task we use a pretrained XLM ~\cite{DBLP:journals/corr/abs-1901-07291} \textit{xlm-mlm-ende-1024}
from Huggingface~\cite{Wolf2019HuggingFacesTS}. BaseNMT+MLM is similar to BaseNMT+LM but it uses {$\log P_{\text{NMT}}(\mbf y|\mbf x) + \lambda \log P_{MLM}(\mbf{y})$}, where  $P_{MLM}$ is the average pseudo-log-probability of sample $\mbf{y}$ calculated using BERT. We use the same architecture of BERT as Marginal-EBM, but we fine-tuned BERT for MLM over the target sentences in training sets for 10 epochs. We tuned $\lambda$ similar to BaseNMT+LM.

EBR is our method that uses rank-based training for EBMs. We explore EBR with Marginal-EBM (Marginal-EBR) and Joint-EBM (Conditional-EBR). We also use noise-contrastive estimation to train our Marginal-EBM, similar to \citeay{deng&al20}, which we refer to as NCE-EBR.  Next, we have Shared-EBR that trains single Marginal-EBM for the tasks with the same target language. Shared-EBR is only trained on IWSLT and FLoRes tasks with English target. For this method, we first sample a translation task and then sample a batch from that task and follow Algorithm~\ref{alg:rank} for the training of the Marginal-EBM.
Finally, as an upper bound for the best achievable result, we also extract the translations from the sample that are closest to the gold data (based on BLEU score).

\subsection{Results} 
Table~\ref{tab:main} shows the performance of the described methods for IWSLT, FLoRes, and WMT translation tasks.\footnote{We use SacreBLEU \cite{post-2018-call} as a consistent BLEU implementation for all of our experiments.} BaseNMT+Sample achieves a better score than beam decoding suggesting that our multinomial sampling supports the modes of the distribution defined by the BaseNMT. Similarly, oracle values are high, indicating that the samples also support the desired distribution. This satisfies the necessary condition for {$P_\theta(\mbf{y}|\mbf{x}) \propto P_{\text{NMT}}(\mbf y|\mbf x)\exp (-E_\theta(\mbf{y},\mbf{x})/T)$} to be closer to the desired distribution. Re-ranking with a language model using BaseNMT+LM improves over BaseNMT+Sample for \ra{De}{En}, \ra{Fr}{En}, \ra{It}{En}, and \ra{En}{De}, but fails on \ra{Ro}{En}, \ra{Si}{En}, and \ra{Ne}{En}. However, in all of these tasks, the difference between BaseNMT+Sample and BaseNMT+LM is not substantial. BaseNMT+MLM is consistently better than BaseNMT+LM. The performance of BaseNMT+MLM is attributable to PLL scoring, as the encoder has the global information over the sentence. Marginal-EBR  performs considerably better than BaseNMT+\{Beam, Sample, LM, MLM\} and better than NCE-EBR on all tasks except on \ra{Ne}{En}, where NCE-EBR outperforms Marginal-EBR. The main advantage of Marginal-EBR over NCE-EBR is the use of only sampled data instead of gold data for training. See Section~\ref{sec:gold} for detailed discussion.

Shared-EBR has a significant improvement over the Marginal-EBR, especially it improves the low-resource task of \ra{Si}{En} by more than 2 BLEU points. 
For this task, we also show that how using more language pairs in training improves performance (Table~\ref{tab:si}). 

Conditional-EBR outperforms Shared-EBR on all tasks. The performance of Conditional-EBR is due to the use of Joint-EBM model, which enables the model to define different energy landscapes for different source sentences. Therefore, samples from the target language are more separable given the source sentence, while Marginal-EBM may not distinguish target sentences for different source sentences.

The translation improvement of using EBR on IWSLT and FLoRes translation tasks are more considerable than the improvement of using EBR on WMT tasks. We believe that pre-trained BERT helps low-resource tasks more than large-scale translation tasks.

\subsection{Effect of Using Gold Data}\label{sec:gold}
Noise-contrastive estimation (NCE) trains the energy model using a discriminative training to distinguish gold data from the sampled data~\cite{gutmann&hyvarinen10,deng&al20}. 
In contrast to the NCE-EBR, EBR does not directly use gold data in the training of the EBM, but only exploit it to determine the rank of two points as well as the margin. 
To show that our approach is effective, we introduce parameter $\gamma$ as the percentage of the time that we can use gold data as one of the points (for example,  $\mbf{y}_h$ in Algorithm~\ref{alg:rank}). Table~\ref{tab:gamma} shows the results for both \ra{De}{En} and \ra{Fr}{En} tasks using Marginal-EBR. As we increase the value of $\gamma$, the performance of Marginal-EBR drops. The main reason is that BaseNMT rarely produces the exact correct translation in the sample set, thus learning the ranking with respect to the gold data is not very informative. When the $\gamma$ is zero, the Marginal-EBM learns to re-rank the samples with respect to their distance to the gold data.

\begin{table}[]
    \centering
    \caption{The effect of using gold data in the ranking objective for Marginal-EBR.}
    \small{
    \begin{tabular}{c c c c c}
    \toprule
        $\gamma$ & 0.0 & 0.25 & 0.75  & 1.0 \\
    \midrule
       \ra{De}{En}  & 35.68 & 35.00 & 34.20 & 33.75\\
       \ra{Fr}{En} & 33.77 & 33.15 & 31.65 & 30.82 \\
    \bottomrule
    \end{tabular}
    }
    
    \label{tab:gamma}
\end{table}
\begin{table}[t]
\centering

\caption{Effect of Entropy Regularization on IWSLT'14 DE-EN}
\vspace{-0.1 in}
\small{
\begin{tabular}{ l c c  } 
\toprule
 &Regularization& No Regularization \\ 
\midrule
BaseNMT + Beam &33.96 & 33.87 \\ 
Conditional-EBR &\textbf{37.88}&37.58 \\
\midrule
Oracle &68.21&67.54 \\
\bottomrule 
\end{tabular}
}
\label{tab:reg}
\vspace{-0.1in}
\end{table}
\subsection{Regularized Training}
We hypothesize that the performance of EBR improves as we increase the support of the base distribution toward the mode of the true distribution. To show that we add an entropy regularization term to the likelihood training of BaseNMT:
\begin{align}
\max_\theta \sum_{(\mbf{x},\mbf{y})\in \mathcal{D}} & \sum_i \log p(y_i| \mbf y_{<i},\mbf x) \nonumber \\
&- \beta \sum_i p(y_i)\log p(y_i).   
\end{align}

Entropy regularization improves the diversity of samples, and as a result, Oracle's score increases by 0.67 BLEU points. While BaseNMT only benefits less than 0.1 BLEU points from the regularization, Conditional-EBR improves by 0.3 BLEU points (see Table~\ref{tab:reg}).
For this study we explored $\beta$ from \{0.01, 0.1\}, and reported results use $\beta=0.01$ selected based on the validation set. BaseNMT trained with $\beta=0.1$ has the Oracle score of 65.76 on the test set (comparing to the Oracle score of 68.21 for $\beta=0.01$), which indicates that stronger regularization reduces the sample quality. 

\subsection{Using XY-BERT for Joint-EBM} \label{sec:xybert}
To explore the effect of a different way of conditioning on the source language, we compare the EBM constructed using the Joint-BERT model with EBM constructed using recently introduced XY-BERT~\cite{guo&al2020}. To construct EBM from XY-BERT, we remove the output layer and project each hidden-state of the final layer to a scalar energy value similar to how we build EBM from BERT. 
We compare these two models on IWSLT'14 \ra{De}{En} task. For XY-BERT we use German BERT for the encoder and English BERT for the decoder, following \citeay{guo&al2020}. Our Joint-BERT uses Multilingual BERT because we feed both source and target sentences to BERT jointly. 
Conditional-EBR with XY-BERT achieves 38.33 BLEU score, which is 0.75 BLEU points higher than Conditional-EBR with Joint-BERT and improves the performance of XY-BERT with mask-predict decoding~\cite{ghazvininejad&al19} by 1.84 BLEU points.\footnote{\citeay{guo&al2020} report 36.49 BLEU score using XY-BERT with 10 iterations of mask-predict decoding.} We believe that the improvement in Conditional-EBR using XY-BERT is mostly attributable to using specialized BERT models. Moreover, XY-BERT has extra trainable modules, so we could fine-tune XY-BERT on the translation task for 60 epochs, while keeping the rest of the parameters fixed without causing catastrophic forgetting. Joint-BERT, on the other hand, does not have any extra parameters, so we fine-tuned all parameters for only 15 epochs. Further training of Joint BERT resulted in poor performance. We leave adding extra modules for better fine-tuning of Joint BERT for future studies. 


\subsection{Maximizing Expected Score}
As another comparison, we train our models by directly maximizing the expected BLEU score (compared to rank-based training):
\begin{align}
\max_\theta \mathbb E_{\mathbf{y}_p\sim p_{ \theta}(.|\mathbf{x})}[\text{BLEU}(\mathbf{y}_p, \mathbf{y}^*)]
\end{align}
We use log-trick to calculate the gradient of the above objective:
\begin{align}
   \nabla_{\theta}& \mathbb{E}_{p_\theta}[\text{BLEU}(\mathbf{y}_p, \mathbf{y}^*)] \nonumber \\
   &= \mathbb{E}_{\mathbf{y}_p\sim p_\theta} [\text{BLEU}(\mathbf{y}_p, \mathbf{y}^*) [-\nabla_\theta E_\theta(\mathbf{y}_p, \mathbf{x}) \nonumber \\ &+ \mathbb E_{\mathbf{y}'\sim p_\theta}[\nabla_\theta E(\mathbf{y}',\mathbf{x})]]].
\end{align}
We use self-normalized importance sampling to draw samples from the energy-based model. We use one sample to approximate the outer expectation and 10 samples to approximate the inner expectation. 
We train both Marginal-EBM and Joint-EBM by maximizing the expected BLEU score on IWSLT'14 DE-EN. The former obtains a score of 34.20 BLEU and the latter achieves 34.77 BLEU points. Both models underperform rank-based training.

\subsection{Inference Time}
We compare the inference latency of EBR variations with BaseNMT (Table \ref{tab:latency}). We use 100 samples for re-ranking using Marginal-EBR, Conditional-EBR with Joint-BERT and Conditional EBR with XY-BERT~\cite{guo&al2020}. Inference on Marginal-EBR takes on average about 170 milliseconds per sentence more than inference in BaseNMT as we have to sample 100 sentences from BaseNMT and evaluate them on the energy model.  
We evaluate the Marginal-EBR only on the target sentences, while we evaluate Conditional-EBR for sequences from both source and target language, so the input sequence of Conditional-EBR is longer, thus having higher latency comparing to Marginal-EBR. We also measure the latency of Conditional-EBR when we use XY-BERT architecture to construct Joint-EBM. In this case, we have two separate BERT models for source and target languages, increasing the number of parameters by 3.3 million and latency by about 90 milliseconds per sentence compared to Conditional-EBR that uses the Joint-BERT model. 

\begin{table}[t]
\centering
\caption{Average inference time per sentence (milliseconds), baseline transformer uses beam width of 5 and EBR uses 100 samples per sentence.}
\vspace{-0.1 in}
\small{
\begin{tabular}{ l c c  } 
\toprule
Method &De$\xrightarrow{}$En& En$ \xrightarrow{}$De \\ 
\midrule
Base-NMT& 572& 577 \\ 
Marginal-EBR& 749& 756 \\ 
Conditional-EBR (Joint BERT) & 836& 838 \\ 
Conditional-EBR (XY-BERT) & 921& 929 \\ 
\bottomrule 
\end{tabular}
}
\label{tab:latency}
\vspace{-0.1in}
\end{table}

\section{Analysis}
In this section, we study the sentence preference of Marginal-EBR created by the energy ranking. 
\subsection{Qualitative Analysis} 
We qualitatively investigate how the output of Marginal-EBR differs from that of BaseNMT model. On the IWSLT'14 test set, we examined 200 examples on which Marginal-EBR did better than NMT and 200 examples where BaseNMT is better. We find that about 30\% of the time, the Marginal-EBR model chooses a translation with changed pronoun. Another frequent `preference' Marginal-EBR makes compared to BaseNMT is to use the contraction form. Since this IWSLT data set is from TED talk, we conjecture that the energy model favors the translations that are in more oral style. Besides, it is also common for the Marginal-EBR model to prefer rephrases, for example, instead of using `will' as used in BaseNMT, Marginal-EBR chooses the form `am going to'. Finally, we find, for some pairs, Marginal-EBR chooses a different tense compared to the BaseNMT model (from MAP decoding). 

Table \ref{tab:analysis-eg} presents quintessential examples we find after examining 400 examples on IWSLT'14 \ra{De}{En} test set. It is worth to mention that examples do not strictly land in only one category. For example, the sentences we show in the `Rephrase` type will also be counted as the change of pronouns. With this in mind, we compute statistics over the 400 sentences and find each of the `Pronoun', `Contraction' and `Rephrase' appears approximately 30\% of the time while 10\% of the sentences change `Tense'. The other less frequent types are changing of determiners, prepositions and deletion (comparing the MAP decoding of BaseNMT and preferred output by Marginal-EBR).

\begin{table}[]
    \centering
    \scalebox{0.75}{\begin{tabular}{cl}
    \toprule
\multicolumn{1}{c}{\textbf{Type}}     & \textbf{Example }                                               \\\hline
\multirow{2}{*}{Pronoun}     & N: to us , \textcolor{blue}{he} meant the freedom .                    \\
                             & E: for us , \textcolor{blue}{it} meant freedom .                     \\\hline
\multirow{2}{*}{Contraction} & N: they are exotic ; \textcolor{blue}{they are} experimental .         \\
                             & E: they are exotical . \textcolor{blue}{they \&apos;re} experimental . \\\hline
\multirow{2}{*}{Rephrase}    & N: and it \&apos;s our \textcolor{blue}{unseen} reality .              \\
                             & E: that \&apos;s our \textcolor{blue}{invisible} reality .           \\\hline
\multirow{2}{*}{Tense}       & N: a new life \textcolor{blue}{has been} born .                        \\
                             & E: and a new life \textcolor{blue}{was} born .  \\ 
                             \bottomrule
\end{tabular}}
    \caption{Typical examples on IWSLT'14 test set, categorized by the difference between BaseNMT and Marginal-EBR. `N' stands for BaseNMT and `E' stands for Marginal-EBR introduced in this paper. }
    \label{tab:analysis-eg}
\end{table}

\subsection{BLEU Gains by Length}

Besides the qualitative analysis, we are also curious to see whether the improvement is affected by length. Table \ref{tab:analysis-length} shows the BLEU scores on the IWSLT'14 test set, which is divided into three bins according to the target length. Shorter sentences have the largest increase in BLEU, and the gain is decreasing as length increases.  We reckon that it is easier for EBR to cover larger training space for sentences of shorter length and thus has the largest improvement in BLEU for these sentences. 

\begin{table}[]
    \centering
        \caption{BLEU scores by length on IWSLT'14 test set. Sentences are divided into 3 groups according to reference length: less than or equal to 5 , in the range between 5 and 10, greater than 10. }
    \small{
    \begin{tabular}{l c c c}
    \toprule
         &  (0, 5] & (5, 10] & (10, )\\
    \midrule
     NMT  & 23.78  & 33.22 & 34.77 \\
     Marginal-EBR & 26.38 & 35.20 & 35.68 \\
     \bottomrule
    \end{tabular}
    }

    \label{tab:analysis-length}
\end{table}

\subsection{Random Sentences}

In the absence of access to the source sentence, the energy model ranks the outputs purely according to the features of target sentences. We hypothesize that the energy model is better at differentiating incoherent and coherent sentences and manage to show that through the following analysis. We apply two kinds of shuffle on IWSLT'14 test set targets: (1) global shuffle: tokens in the sentence are randomly shuffled (2) local shuffle: we first randomly select a token and randomly shuffle the tokens within a local window of three. Then we compute the energy scores of these shuffled sentences as well as the untouched ones. The energy scores are listed in Table~\ref{tab:analysis-random}. (The energy model assign a lower energy to its preference.) We find 87\% of the time, the energy model is able to distinguish the original sentence from a local shuffled one, and 90.5\% from the global shuffled one. This supports our hypothesis that the energy model is capable of capturing the fluency of generated candidates.

\begin{table}[]
    \centering
       \caption{Energy scores of randomly shuffled sentences as well as original targets on IWSLT'14 \ra{De}{En} test set.}
    \small{
    \begin{tabular}{cc}
    \toprule
        Shuffle Type & Average Energy Scores \\ \hline
        Local & -0.013 \\
        Global & 0.002 \\
        Original & -0.037 \\
        \bottomrule
    \end{tabular}
    }
 
    \label{tab:analysis-random}
\end{table}



\section{Conclusion and Future Work}
We introduce energy-based re-ranking (EBR) to improve the performance of autoregressive neural machine translation.
Despite its superior performance, EBR suffers from high latency because of its dependency on sampling from an autoregressive model. Directly sampling from the underlying EBM can speed up the inference, which is our future direction in order to benefit from the power of energy-based models for machine translation.


\bibliography{all}
\bibliographystyle{acl_natbib}

\end{document}